\documentclass{article}

     \usepackage[final,nonatbib]{neurips_2019}

\usepackage[utf8]{inputenc} % allow utf-8 input
\usepackage[T1]{fontenc}    % use 8-bit T1 fonts
\usepackage{hyperref}       % hyperlinks
\usepackage{url}            % simple URL typesetting
\usepackage{booktabs}       % professional-quality tables
\usepackage{amsfonts}       % blackboard math symbols
\usepackage{nicefrac}       % compact symbols for 1/2, etc.
\usepackage{microtype}      % microtypography

\usepackage[ruled,vlined,linesnumbered]{algorithm2e}
\usepackage{subfigure}
\usepackage{color,xcolor}
\usepackage{graphicx}
\usepackage[toc]{appendix}

\title{Object Localization with a Weakly Supervised CapsNet}

\author{%
  Weitang Liu\\
  Department of Electrical and Computer Engineering\\
  University of California, Davis\\
  Davis, CA 95616 USA \\
  \texttt{wetliu@ucdavis.edu} \\
  \And
  Emad Barsoum\\
  Microsoft\\
  One Microsoft Way, WA 98052 USA\\
  \texttt{ebarsoum@microsoft.com} \\
  \AND
  John D. Owens\\
  Department of Electrical and Computer Engineering\\
  University of California, Davis\\
  Davis, CA 95616 USA\\
  \texttt{jowens@ece.ucdavis.edu} \\
}

\begin{document}

\maketitle

\begin{abstract}
   Inspired by CapsNet's routing-by-agreement mechanism with its ability to learn object properties, we explore if those properties in turn can determine new properties of the objects, such as the locations. We then propose a CapsNet architecture with object coordinate atoms and a modified routing-by-agreement algorithm with unevenly distributed initial routing probabilities. The model is based on CapsNet but uses a routing algorithm to find the objects' approximate positions in the image coordinate system. We also discussed how to derive the property of translation through coordinate atoms and we show the importance of sparse representation. We train our model on  the single moving MNIST dataset with class labels. Our model can learn and derive the coordinates of the digits better than its convolution counterpart that lacks a routing-by-agreement algorithm, and can also perform well when testing on the multi-digit moving MNIST and KTH datasets. The results show our method reaches the state-of-art performance on object localization without any extra localization techniques and modules as in prior work. 
\end{abstract}

\section{Introduction}
Humans in their early years learn the motion of objects intuitively. They are able to not only learn to recognize objects, but also locate objects and predict their motion. They gradually learn to summarize the motion rules that apply to objects they have never seen before. In kinematics problems, such as predicting the trajectory of a ball, objects are often simplified as point masses~\cite{whittaker1988treatise}, which serve as the input to equations of motion. This demonstrates coordinates are one of the key concepts that unify solutions with different objects in kinematic problems. 

In physical simulation frameworks, the physics of the environment is modeled by predefined rules. These rules determine how objects move in the scene and the graphics system utilizes the affine transformation matrices generated from the physics system to render objects on the screen. Our idea is to reverse this process. As a powerful function approximation tool, the neural network can determine the affine transformation matrices for each object, frame-by-frame, through the learned physics rules of the underlying physical environment. Such a network should be able to model the rules that generate transformation matrices not only for objects it has seen before, but also for unknown objects in the same environment. This is similar to the case of predicting the trajectory of a baseball through learning the trajectory of throwing other objects. In this paper, we focus on deriving the objects' coordinates which are important in trajectory prediction. 

As a variant of a convolution neural network (CNN), CapsNet~\cite{sabour2017dynamic} has been recently proposed to better describe objects' properties with vector-based activation and specially designed routing algorithms. The capsules encode the pose information that is more general than a scalar value representing the presence of an object as in CNN. The routing algorithm makes this pose information useful to form object hierarchies. Low-level capsules activate and vote by pose agreement to determine which high-level capsules will be activated, creating a part-whole relationship between the low-level capsule and the high-level capsules. As a result, each dimension of a capsule represents some properties of the object, such as stroke thickness of digits in the MNIST dataset. Moreover, by manually changing the value of each dimension in the digit capsules, Sabour et al.~\cite{sabour2017dynamic} showed that these properties can be controlled and visualized through reconstruction layers. As low-level capsules that share similar poses are routed to form high-level capsules, coordinates of high-level capsules' centers can be derived by averaging the activated low-level capsules. These high-level coordinates will be able to approximate the position of objects' centers.

Our contributions are as follows. First, we propose an algorithm to derive the approximate coordinates of object centers in the image coordinate system through CapsNet (Sec.~\ref{sec:caps}). The importance of this algorithm is that it shows routing-by-agree algorithm not only discovers the part-whole relationships, but those relations in turn can derive new properties of the objects. Second, we show the importance of sparse capsule representation (Sec.~\ref{sec:sparse}). This shows the importance of bias toward the "unknown" capsule instead of having equal uncertainties of the capsules of next layer. Third, following this idea, we propose a bias routing probability initializer to preserve the capsules that substantially contribute to the classification and reconstruction, a technique which helps improve the weakly supervised object localization task. Finally, we demonstrate through ablation studies to show how the bias initializer influences the performance of classification and weakly coordinate derivation (Appendix.~\ref{A:ablation}). 

\section{Related Work \label{sec:related}}
Sabour et al.~\cite{sabour2017dynamic} recently proposed CapsNet with a dynamic routing algorithm. By replacing the max-pooling layer with a routing-by-agreement algorithm, CapsNet ensures that lower-level capsules send their information to higher-level capsules in order to achieve equivariance. By explicitly concatenating the coordinates of the lower-level capsules and then routing-by-agreement, our proposed algorithm can derive the objects' center of mass at the higher-level capsules from the lower-level capsules. Recently, Hinton et al.~\cite{46653} proposed the EM (expectation maximization) routing algorithm with a coordinate addition. Our algorithm is inspired by the coordinate addition, but we explicitly utilize this coordinate information in the reconstruction layers. We suggest that the coordinates are useful not only in routing as Hinton et al.\ mentioned, but also in deriving the coordinates of high-level capsules as a fundamental property in learning translational motions. 

Many algorithms identify the positions of objects. Supervised learning algorithms~\cite{liu2016ssd,redmon2016you,ren2015faster} require bounding boxes and classes as inputs to train the recognition networks. Weakly supervised methods~\cite{selvaraju2016grad, zhou2018weakly} usually only require the objects' classes as inputs to train the network. However, the objects' positions are derived from active spots represented by the gradients of the last few convolution layers or Peak Response Maps modules. Other methods~\cite{Novotny_2018_ECCV,frohlich2012semantic} focus on segmentation of the image in a supervised learning setting. Our method focuses on deriving the coordinates as a by-product of the CapsNet without any extra modules; it requires the input image, classification ground truth, and coordinate addition in the routing process. By manually tweaking the capsules' coordinate atoms, the reconstruction layers will generate images with the object appear in the modified position. This shows the coordinates our method finds are controllable properties of the objects.

Srivastava et al.~\cite{srivastava2015unsupervised} has proposed an LSTM model for predicting future frames on the moving MNIST dataset with two digits and on the UCF-101 dataset, which is more complicated than our moving MNIST dataset. The future frame predictions of the work are purely unsupervised. Our proposed method focuses on the fact that CapsNet, once it is well-trained, is able to reason out object properties, such as the coordinates of objects' centers, that are potentially useful for generalization and transfer learning. 

Other related work includes spatial transformer networks~\cite{jaderberg2015spatial} injects a spatial transformation module to improve classification performance. Our work instead utilizes the agreement between lower-level capsules to find equivariance. Moreover, Transforming Autoencoder~\cite{hinton2011transforming} defines capsules and how they should work, while Sabour et al.~\cite{sabour2017dynamic} proposes a deep routing-by-agreement algorithm, which focuses on how to utilize the pose agreement phenomenon. 

%------------------------------------------------------------------------
\section{Model Architecture}
\subsection{CapsNet Preliminary}
CapsNet~\cite{sabour2017dynamic} proposes a routing-by-agreement algorithm to achieve equivariance. An advantage of the CapsNet over CNN is that CapsNet encourages the equivariance, such as pose agreement of an entity and its parts. 

CapsNet consists of three major modules, convolution layers for simple feature extraction like other CNN, a primary capsule layer, and a digit capsule layer. After the convolution layers, it splits the channel axis into several vector representations called capsules, which preserve the pose representations of the objects (or parts of the objects). This layer is called the primary capsule layer.  Afterward, each primary capsule $i$ is transformed through some transformation matrices given class $j$ into a set of capsules $\bold{\hat{u}}_{j|i}$, activated through squash function, and routed with the other capsules. If the transformed lower level capsules (primary capsules in this case) tend to cluster together, the routing-by-agreement algorithm finds the center of the clusters as the next level capsule (digit capsule). Finally, reconstruction layers generate the input image, working as a regularizer. In CapsNet's setting, the classification loss function is the difference between the norm of the digit capsules and the ground truth class label.

Primary capsule $i$ (transformed given the digit capsule $j$), with atoms $\bold{\hat{u}}_{i|j}$, has a certain logit $b_{ij}$ and probability $c_{ij}$ of contributing to that digit capsule $j$. For every iteration, it checks whether the next digit cap $j$ is activated and what direction it is $\bold{v}_j$, and increases the probability $c_{ij}$ for the primary capsules that agree with the direction of $\bold{v}_j$ given that capsule $j$ is activated by the $\textit{squash}$ function.

\subsection{CapsNet with coordinate atoms \label{sec:caps}}
We adopt a similar architecture to CapsNet~\cite{sabour2017dynamic}. After the primary-caps layer, the capsules are multiplied by the transformation matrices. We then concatenate two additional atoms to each of the capsules, according to the capsules' relative position in the feature map. Thus, the capsules are 18 dimensions instead of 16. These two atoms are normalized from 0 to 1. This setting is similar to the coordinate addition Hinton et al.\ performed in the EM-Routing version of CapsNet~\cite{46653}. For example, if the feature map after the primary-caps layer has spatial dimensions of $7\times7$, and the calculated receptive field is $9\times9$, and thus the radius of the receptive field is 4. The capsule representing the top-left corner has coordinate values of $[x,y] = [\frac{4}{64}, \frac{4}{64}]$, where 64 is the original image size and 4 is radius of the receptive field. Different receptive fields have overlapping areas. The routing algorithm also considers these two coordinate atoms as well. During the routing, 16 atoms, excluding the two coordinate atoms, are passed to the activation function. To simplify the routing process, we concatenate the coordinate atoms and the other 16 atoms only when the logits are updated and output capsules are returned. The routing probabilities of the low-level capsules, $c_{i}$, are utilized to determine the coordinates of the high-level capsules through a weighted average: 

\begin{equation} \label{equ:1}
\textit{\textbf{coordinates}}_{j}^{l} = \frac{\sum_{i}^{n}c_{i,j}\textit{\textbf{coordinates}}_{i,j}^{l-1}}{\sum_{i}^{n}c_{i,j}}
\end{equation}

\noindent
where the $l$ denotes the digit-caps layer, and $l-1$ denotes the primary-caps layer. The $n$ denotes the number of in channels, and $j$ denotes the $j$th output capsule. The $\textit{\textbf{coordinates}}_{ij}^{l-1}$ only depends on the spatial position $i$ of the feature map. This equation simply takes the object coordinates from each lower-level capsule and weights them by the coupling coefficient, which predicts how closely related the lower-level and higher-level capsules are. The $\textit{\textbf{coordinates}}$ represents a vector of the coordinates. 

\begin{figure}
  \centering
  \includegraphics[width=0.7\textwidth]{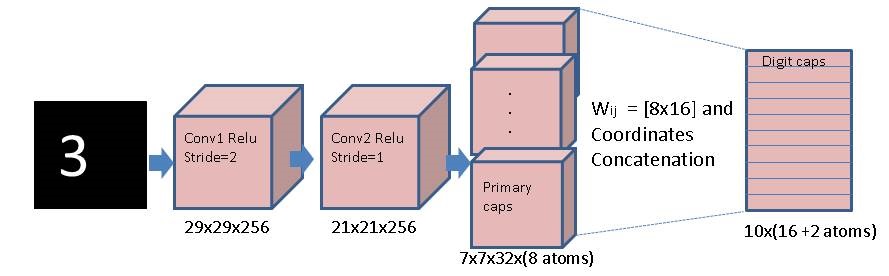}
  \caption{\label{fig:encoder} Modified CapsNet layers. Two convolution layers with different strides extract features before the primary-caps layer. In the primary-caps layer, each capsule after being multiplied by the weight matrix is then concatenated with the coordinate atoms before routing.}
\end{figure}

\begin{figure}
\centering
\includegraphics[width=0.7\textwidth]{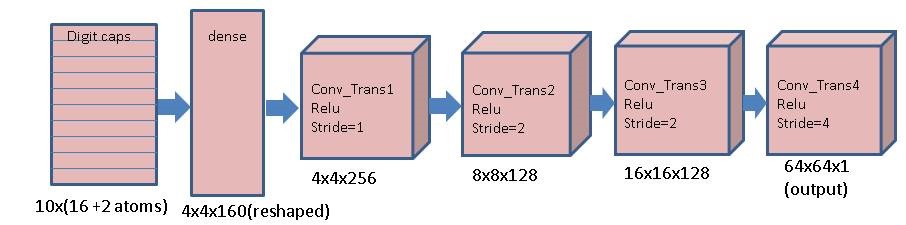}
\caption{\label{fig:decoder} Modified reconstruction layers. The digit-caps layer is flattened and fed as an input to a fully-connected layer. Convolution transpose layers are used. The output of the last convolution layer is a 64$\times$64 image.}
\end{figure}

The new architecture for the CapsNet is summarized in \autoref{fig:encoder}, and the reconstruction layers are shown in \autoref{fig:decoder}. The figures shown here are in HWC format, the input image is the number 3, and 10 is the number of outputs representing the 10 digits. The CapsNet has two convolution layers, each of which has 256 9$\times$9 filters. The stride of the first convolution layer is 2, while the stride of the second convolution layer is 1. A primary-caps layer with a stride equals to 2 has 32 9$\times$9 kernels with 8 atoms. Each capsule with 8 atoms is then multiplied by weight matrices $W_{ij}$ to get 16 atoms. Each capsule is then concatenated with two coordinate atoms before routing. The total number of atoms in a capsule is then 18, thus the size of the digit-caps layer (after routing) is 10$\times$18. Please note that the new coordinate atoms also participate in the routing process, encouraging the neighbor capsules to group together. In order to reduce the number of parameters and preserve the spatial features for the localization, we build the reconstruction layers as convolution-transpose layers instead of fully connected layers. Our reconstruction layers are four convolution layers with 3$\times$3, 3$\times$3, 4$\times$4, 8$\times$8 kernels. Both the classification loss and reconstruction loss are the same as in CapsNet~\cite{sabour2017dynamic}. The routing algorithm with the coordinates is described by \autoref{algo:coordrout}.

\begin{algorithm}
\SetAlgoLined
\KwResult{Digit-caps with coordinates }
 for capsules $i$ in primary-caps layer and $j$ in digit-caps layer: initialize bias logits $b_{ij}$ \;
 for all capsules $j$ in digit-caps: $\bold{\hat{u}}^{\small{\textit{new}}}_{j|i} = \textit{concatenate}(\textit{\textbf{coords}}_{i,j}, \bold{\hat{u}}_{j|i})$ \;
 \For{r iterations} {
  for all capsules $i$ in primary-caps: $\bold{c}_{i} = \textit{softmax}(\bold{b}_{i})$\;
  for all capsules $j$ in digit-caps: $\bold{s}_j = \sum_i c_{ij}\bold{\hat{u}}_{j|i}$\;
  for all capsules $j$ in digit-caps: $\bold{v}_j = \textit{squash}({\bold{s}}_{j})$\;
  for all capsules $j$ in digit-caps: $\bold{o}_j = \frac{\sum_i c_{ij}\textit{\textbf{coords}}_{ij}}{\sum_i c_{ij}}$ \;
  for all capsules $j$ in digit-caps: $\bold{v}^\textit{new}_j = \textit{concatenate}(\bold{o}_j, \bold{v}_j)$ \;
  for all capsules $i$ in primary-caps and for all capsules $j$ in digit-caps: $b_{ij} = b_{ij} + \hat{\bold{u}}^{\small{\textit{new}}}_{j|i}\cdot \bold{v}^{\small{\textit{new}}}_j$\;
 }
 \caption{Routing with coordinates from primary-caps to digit-caps. Line 2, 7 and 8 are our contributions. We discuss the bias initializer in~\autoref{sec:sparse}. \label{algo:coordrout}}
\end{algorithm}

\subsection{Routing-by-agreement algorithm as sparse capsule learner \label{sec:sparse}}
We notice that the routing-by-agreement algorithm can generate sparse routing probabilities $c_{ij}$, which are essential in our localization task. In the original implementation of the routing-by-agreement algorithm, there is a leaky-routing component that works as "unknown" class. For example, if there are 10 classes as ground truth, there will be 11 digit-capsules. The initial probabilities $c_{ij}$ before routing are 0.09 (1/11). This leaky-routing gives the network the ability to route the capsules in the primary-caps layer to this "unknown" class, instead of forcing each capsule to make decision to a specific class. 

From our preliminary experiment, we found that this evenly distributed $c_{ij}$ of the original implementation work well in classification. However, we also found that almost all the routing probabilities $c_{ij}$ after routing have similar values. Thus, most of the capsules in the primary-caps layer almost contribute equally and significantly to the digit capsule. Figure~\ref{fig:eveninit} shows a random sample's final routing probabilities $c_{i}$ of each capsule in the primary-caps layer to the ground truth digit-capsule $j$ after the CapsNet converges. Most of the capsules have routing probability around 0.1, while the highest routing probability is around 0.16. The fact that the range of the routing probabilities is small leads to a problematic result in \autoref{algo:coordrout}, because the weighted average of the coordinates in equation~\ref{equ:1} will converge to a constant that misses the information of the capsules with high routing probabilities. 

A simple reasoning of the conclusion is as follows. Suppose that there are $n_0$ capsules with routing probabilities $c_{ij}=c_{0}$, and $n_1$ capsules with routing probabilities $c_{ij} = c_{1}$, where $c_0$ and $c_1$ are two constant values. Equation~\ref{equ:1} will become 
\begin{equation}
\textbf{\textit{coords}}_{j}^{l} = \frac{\sum_{i}^{n_0}c_{0}\textit{\textbf{coords}}_{ij}^{l-1}+\sum_{i}^{n_1}c_{1}\textit{\textbf{coords}}_{ij}^{l-1}}{\sum_{i}^{n_0}c_{0} + \sum_{i}^{n_1}c_{1}}
\end{equation}
In the case that $n_0 >> n_1$, and $c_1\approx c_0$,
\begin{equation} \label{equ:2}
\textit{\textbf{coords}}_{j}^{l} \approx \frac{\sum_{i}^{n_0}c_{0}\textit{\textbf{coords}}_{ij}^{l-1}}{\sum_{i}^{n_0}c_{0}}
\end{equation}
However, only the capsules that have higher routing probabilities are contributing to the classification, and thus they should be weighted more than the average capsules instead of being averaged-out as in equation~\ref{equ:2}.

In order to derive the coordinates of the objects, substantial differences in the routing probability $c_{ij}$ between the capsules are needed. Following the idea that the primary capsules should know nothing about the digit capsules, we initialize the $c_{ij}$ to a very small probability if $j$ is the digit class, and to a very high probability if $j$ is an unknown class. In this setting, many of the primary capsules will not be assigned a high probability after the routing-by-agreement algorithm, unless the primary capsules really contribute to the classification to the digits. In the end, only a few primary capsules are activated and assigned high probability $c_{ij}$ after routing, and these capsules can serve the object localization task well. 

\section{Experiments}
\subsection{Datasets and benchmark models}
These models are tested using the moving MNIST dataset and the moving Fashion MNIST dataset generated through Szeto et al.~\cite{szeto2018dataset}, which can set different moving speeds of the digits. The speed of each sample is a random variable ranging from 1 to 9. The samples have translation motion only. The sequence length is 20, which is the number of frames per sample (time axis). Each sample contains only one handwritten digit, generated by placing a MNIST sample on a 64$\times$64 canvas. We generate two training sets and two test sets for moving MNIST and moving Fashion MNIST; each training set has 30,000 samples and each test set has 1,000 samples. 

In comparison, we provide two benchmark models, where we remove all the routing-by-agreement structures, replace the primary-caps layer with a convolution layer, and the digit-caps layer with the fully-connected layer. The first benchmark model has the same structure as the number of output neurons as the digit cap layer. This means that a 10x16 matrix is an output from the fully-connected layer after reshape, where 10 is the number of classes, and 16 is the embedding dimension. The norms of this matrix over the rows are taken to generate a 10x1 vector, which serves as the logits to a softmax classifier. Similar to our CapsNet, this matrix is the input to reconstruction layers. The second benchmark model only has 10 neurons as the logits to softmax after the fully-connected layer.

For the benchmark models, we adopt the method from GradCam~\cite{selvaraju2016grad} and the TensorFlow implementation\footnote{https://github.com/Ankush96/grad-cam.tensorflow} of Gracam, but use a weighted mean based on the gradient map for each image. We discussed why we did not choose other benchmark models in Appendix~\ref{A:benchmark}. The $x$ coordinate of the object is: 
\begin{equation}
\textbf{x}^\textit{coord} = \frac{\sum_{i,j} \textbf{x}^\textit{coord}_{ij} * \textit{gradmap}_{ij}}{\sum_{i,j} \textit{gradmap}_{ij}} 
\end{equation}
where $x^\textit{coord}_{ij}$ is the x coordinates of the ``pixel'' in the final feature map, which is 7x7 in this test. The $\bold{y}^\textit{coord}$ is calculated similarly. Moreover, we also add a heuristic mean absolute error for reference, which is the absolute difference between coordinates (0.5,0.5) and the ground truth trajectory. In this heuristic, we simply assume that the object(s) is(are) always in the middle position of the image.

\subsection{CapsNet as a position identifier\label{sec:exp_caps}}
The first experiment tests if CapsNet is able to extract the coordinates within the image coordinate system. Thus, only images are considered as samples. We randomly pick and shuffle 3 frames out of the 20 frames in the time axis. The result is the number of samples in this experiment: $3\times 30,000=90,000$. The reconstruction balanced factor is 0.005. The number of routing iterations is 5 instead of 3. Our test cases contain three different datasets, each of which contains a different number of digits on the canvas. The ground truth coordinates (trajectory) are provided by datasets but are not used in the training process.

We build our model according to the architecture in \autoref{sec:caps}. The batch size is 100; the number of training epochs is 150; the initial $c_{ij}$ in \autoref{sec:sparse} for the "unknown" class is 99.75\%. The $b_{ij}$ for the "unknown" class is $1+\log(10)$, and the $b_{ij}$ for the real classes are $-5$. The first model we compare is the model with the same architecture but with an initial $c_{ij} = 9\%$ for all classes, where the logits $b_{ij}$ are all 0, a setting that is the same as the original CapsNet. 

\begin{figure}
\centering
\subfigure[]{\includegraphics[width=13mm]{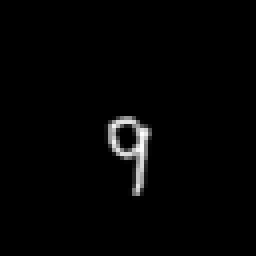}}
\subfigure[]{\includegraphics[width=13mm]{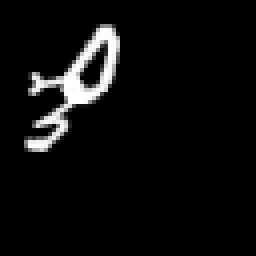}}
\subfigure[]{\includegraphics[width=13mm]{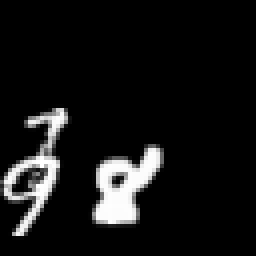}}
\subfigure[]{\includegraphics[width=13mm]{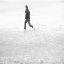}}
\subfigure[]{\includegraphics[width=13mm]{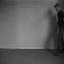}}
\subfigure[]{\includegraphics[width=13mm]{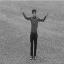}}
\caption{Examples of images in the MNIST (a-c) and KTH (d-f) test sets for localization tasks. \label{fig:ex}}
\end{figure}

Lastly, we also test our model's performance on a multi-digit MNIST dataset, though we only train our models and benchmark models on a single-digit MNIST dataset. The multi-digit test ensures that the models cannot predict the coordinates simply by averaging the white pixels of the image or averaging the active neurons in the convolution feature maps. They have to learn to distinguish the digits and predict different coordinates for each of the digits on the canvas. Limited by the current setting of CapsNet, multi-digits with the same class are not allowed. For CapsNet, the inputs are the image with multiple digits and the binary vectors of ground truth labels. For example, in a sample with three digits, 1 and 5 and 7, the vector is [0,1,0,0,0,1,0,1,0,0]. For the benchmark models, our test will loop over each label of the ground truth to get the individual gradient maps for each digit on the image. Examples of images of 1-, 2-, and 3-digit moving MNIST datasets are shown in figure~\ref{fig:ex}. The digits may overlap with each other. We scale the coordinates (x,y) of an image: the top left corner of the image is (0,0), the top right corner is (0,1), the bottom left corner is (1,0), and the bottom right corner is (1,1). 

These four models, including the benchmark models, are trained on the 1-digit moving MNIST dataset. We only use classification loss (and reconstruction loss if the model reconstructs the input image) to train the models. The ground truth trajectory is not available throughout training stage. 

\begin{figure}
\begin{tabular}{lcccccc}
Frames for digit 3&
\includegraphics[width=.07\linewidth]{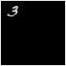}
\includegraphics[width=.07\linewidth]{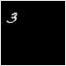}
\includegraphics[width=.07\linewidth]{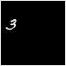}
\includegraphics[width=.07\linewidth]{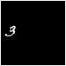}
\includegraphics[width=.07\linewidth]{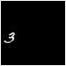}
\includegraphics[width=.07\linewidth]{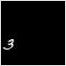}
\\
x atom & 
\small{0.11 \ \ \quad 0.20 \ \ \quad 0.34 \ \quad 0.48 \ \quad 0.64 \ \quad 0.78} \\
y atom & 
\small{0.14 \ \ \quad 0.14 \ \ \quad 0.13 \ \quad  0.11 \ \quad  0.10 \ \quad  0.10} \\
\\
\end{tabular}
    \caption{\label{fig:examples}Coordinate atoms of 6 consecutive ground truth frames for digit 3. The size of the digit is resized to 16$\times$16 for demonstration purposes. The coordinate atoms are normalized from 0 to 1. 0 represents the top (x atom) or left (y atom) of the canvas, while 1 represents the bottom (x atom) or right (y atom) of the canvas.}
\end{figure}

\begin{table*}[t]
 \centering
 \begin{tabular}{ |p{4.5cm}||p{2.1cm}|p{2.1cm}|p{2.1cm}|  }
  \hline
  \multicolumn{4}{|c|}{Multi-digits moving MNIST} \\
  \hline
  Models & MNIST 1 digit & MNIST 2 digit & MNIST 3 digit\\
  \hline
  Heuristic coordinates                  & 0.19  & 0.14  & 0.19 \\
  ResNet 20 w/o reconstruct~\cite{Lin:2018:BDL} & 0.049 & 0.070 & 0.213 \\
  ConvNet w/o reconstruct                & 0.057 & 0.135 & 0.160 \\
  ConvNet w/ reconstruct                 & 0.044 & 0.071 & 0.200 \\
  CapsNet w/ the same initial $c_{ij}$   & 0.192 & 0.196 & 0.187 \\
  \hline
  CapsNet w/ bias initial $c_{ij}$(ours) & 0.038 & 0.053 &  0.078 \\
  \hline
 \end{tabular}
 \caption{Mean absolute difference between ground truth coordinate and predicted coordinates of the four models. The coordinates are scaled to 0 to 1 horizontally and vertically.}
 \label{tab:1}
\end{table*}

Figure~\ref{fig:examples} shows the values of the $x$-$y$ coordinate atoms derived from our model. \autoref{tab:1} shows the mean absolute difference between the ground truth coordinate and predicted coordinates of the CapsNets and their convolution counterpart models (benchmarks). The reconstruction layer serves as a regularization of the model, and our model with the proposed coordinates extraction algorithm and the bias initializer of the routing probabilities $c_{ij}$ can perform better than the coordinates derived from all the other models in this weakly supervised learning setting, where the model only learns to classify instead of learning to predict the coordinates directly. Moving MNIST with 2 and 3 digits are not in the training set, and yet our model is still able to distinguish and localize the digits better than other models. Our model performs 13\%, 24\%, and 51\% better for coordinate derivation on the 1-, 2-, and 3-digit moving MNIST dataset than the best of the other models. 

\begin{figure} 
\centering
\subfigure[Final routing probabilities for each capsule with bias initializer. \label{fig:biasinit}]{\includegraphics[width=60mm]{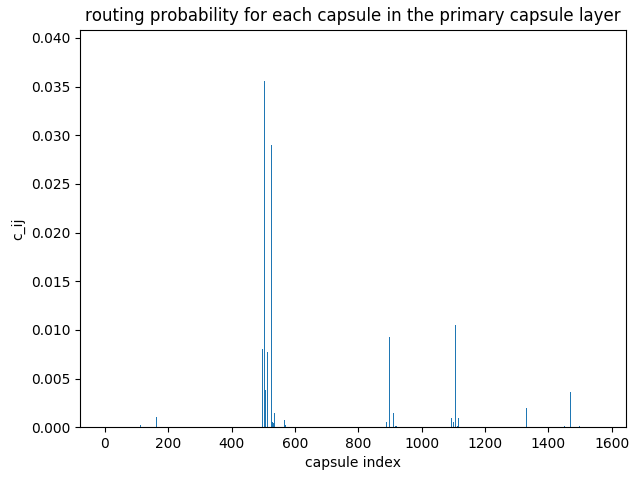}}
\subfigure[Final routing probabilities for each capsule with even initializer.\label{fig:eveninit}]{\includegraphics[width=60mm]{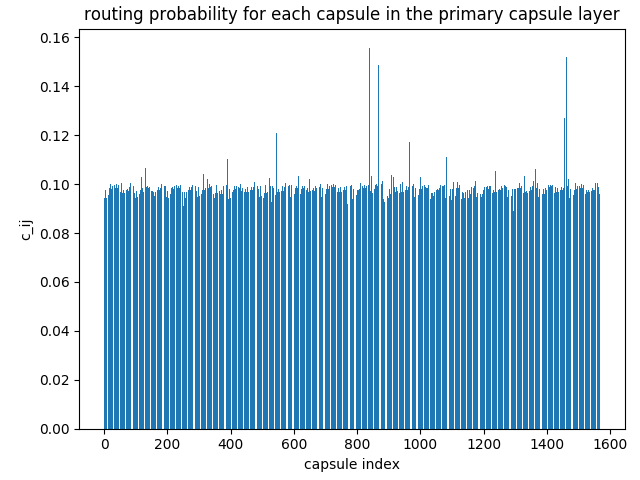}}
\caption{Probabilities $c_{i}$ of routing to the ground truth digit capsule $j$ for each primary capsule. One sub-graph is shown for one image sample.}
\end{figure}

Figure~\ref{fig:biasinit} and~\ref{fig:eveninit} are routing probabilities for two different samples. Our proposed bias initializer, as shown in figure~\ref{fig:biasinit} produces substantial differences in probabilities between capsules. Only the very few capsules that are really contributing to the classification have relatively high routing probabilities. 

Finally, five digits are generated by manipulating the coordinate atoms of the learned capsules as shown in figure~\ref{fig:manipulate}. Each digit has three rows for demonstration. The first row is for x-axis (top-down), and the second row is for y-axis (left-right). The third row is for reference. It is clear that the coordinate atoms can control what the reconstruction layers generate; it works as we expect with the coordinate settings. If the values of the (manipulated) coordinate atoms are larger are smaller than 0 or larger than 1, it will generate weird looking images because we normalize the values of these two atoms from 0 to 1. 

\subsection{CapsNet as a position identifier for KTH dataset ~\label{sec:exp_kth}}
We also test our model on the KTH dataset, which is a human action recognition dataset with video frames. A person in each sample takes a certain action. The ground truth labels are walking, jogging, running, boxing, hand waving, and hand clapping. The people appearing on the images with the first three labels will move to different locations. We preprocess the KTH dataset as in Villegas et al.~\cite{villegas2017decomposing}. From the data list, we generate 37,960 training images, and manually label extra 100 images to get the coordinates of the person in the image. We resize the images of the dataset to 64$\times$64. The coordinates setting is the same as the moving MNIST dataset. Examples of the KTH dataset we use are in figure~\ref{fig:ex}.

Training on KTH is a relatively more difficult task. Compared with the MNIST datasets, KTH has a different goal which is action recognition. Some classes with a person moving out of the camera scenes, such as walking, generate less useful frames than classes such as handwaving where a person doesn't move. We clip 20 frames for all video samples, with only 37,960 training samples (KTH) vs 90,000 (MNIST). As a regularizer, the reconstruction layers improve the performance on MNIST, but take many accounts for the background and shadow of a person on KTH\@. We then modified our model by changing the convolution kernels spatial sizes from 9$\times$9 to 4$\times$4 with stride 2 and the convolution kernels in the primary capsule layer to 2$\times$2 with stride 2. We also set the bias logits $b_{ij}=-7.0$ for the real classes.

We set the training epochs to 200, where the model achieves the lowest mean absolute coordinates difference between the predicted coordinates and the ground truth coordinates. The test accuracy for this test setting is around 74\% to 82\% for different trials. We also add similar heuristics for KTH as well. Table~\ref{tab:2} summarizes the results. Our model can perform slightly better than ResNet-20 benchmark through GradCam.

\begin{figure}
\centering

\end{figure}

\begin{table}
 \centering
 \begin{tabular}{ |p{4.5cm}||p{3cm}|}
  \hline
  \multicolumn{2}{|c|}{KTH dataset} \\
  \hline
  Models                                 & mean absolute error \\
  \hline
  Heuristic coordinates (0.5,0.5)        & 0.113 \\
  ResNet-20                              & 0.090 \\
  CapsNet w/ bias initial $c_{ij}$(ours) & 0.088 \\
  \hline
 \end{tabular}
 \caption{Mean absolute difference between ground truth coordinate and predicted coordinates of the models for KTH dataset.}
 \label{tab:2}
\end{table}

Finally, as we did for the MNIST dataset, we generate four random samples of KTH by manipulating the coordinate atoms of the learned capsules as shown in fig~\ref{fig:manipulate_kth} in Appendix~\ref{A:manipulation}. Each class has three consecutive rows for demonstration. The first row is for x-axis (top-down), and the second row is for y-axis (left-right). The third row, which is generated by manipulating a non-coordinate atom, is for reference. Compared to the MNIST, which has a large amount of samples, the generated KTH images are blurry and the manipulation effect is less clear. However, except for the first class which generates blurry images (first three rows), each of the other classes still has clear manipulation effect on the left-right axis (rows number 5, 8, 11) compared to the reference rows (row number 6, 9, 12). The effect of top-down manuplation is less obvious because the top-down movement in the KTH dataset is relatively rare, while the left-right movement in classes such as running, walking.  

\section{Discussion and future work}
In this paper, we showed a novel use of CapsNet to both predict a translation transform for a moving object, and transfer such knowledge to an unknown object without retraining. One of the main characteristics of our CapsNet is to derive object spatial information. We believe this approach is promising, but we note some limitations, such as the inability of generating a proper orthogonal representation when the atom coordinate represents stroke thickness in addition to the translation coordinate.
\pagebreak

%\small
%\bibliographystyle{plain}
%\bibliography{egbib}

\appendix
\section{Manipulation of the capsule embeddings}\label{A:manipulation}
\begin{figure}
\centering
\includegraphics[width=0.45\textwidth]{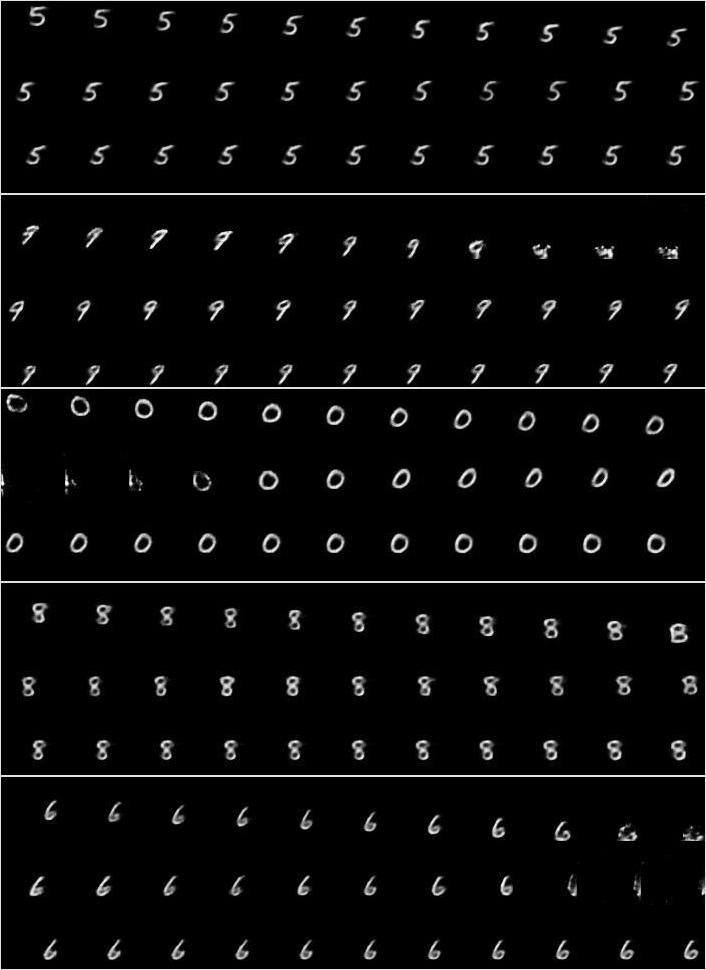}
\caption{\label{fig:manipulate} Manipulating the coordinates of the learned capsules. Each row is generated by first deriving the coordinate atoms and then linearly manipulate its value from -0.20 to +0.20.}
\end{figure}

\begin{figure}
\centering
\includegraphics[width=0.45\textwidth]{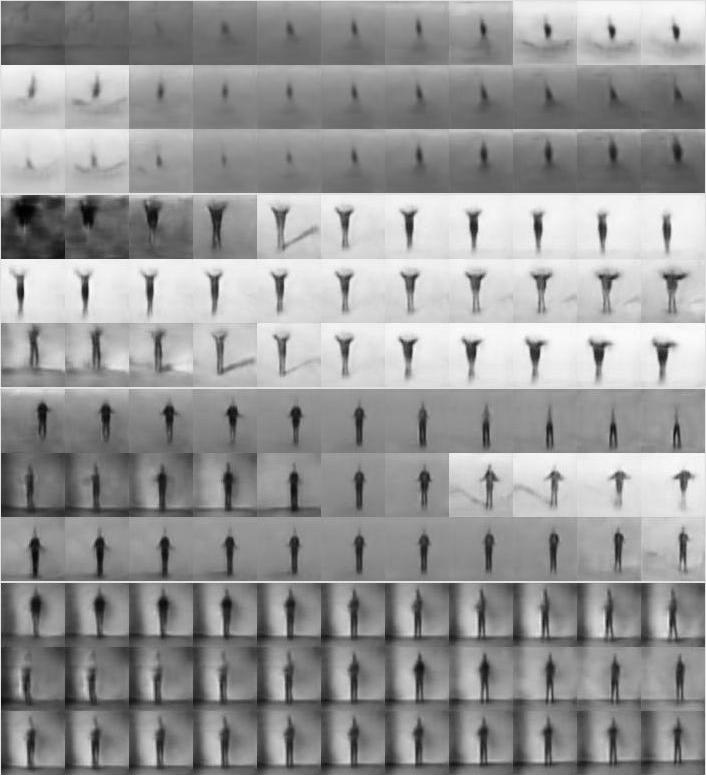}
\caption{\label{fig:manipulate_kth} Manipulating the coordinates of the learned capsules. Each class has three consecutive rows for demonstration. Each row is generated by first deriving the coordinate atoms and then linearly manipulate its value from -0.20 to +0.20. Rows 1, 4, 7, 10 are generated by manipulating the x atoms (top-down). Rows 2, 5, 8, 11 are genereated by manipulating the y atoms (left-right). The reference rows 3, 6, 9, 12.}
\end{figure}

See fig~\ref{fig:manipulate} and fig~\ref{fig:manipulate_kth}.

\section{Ablation studies}\label{A:ablation}
\begin{figure}
\centering
\subfigure[\label{fig:abla1}]{\includegraphics[width=80mm]{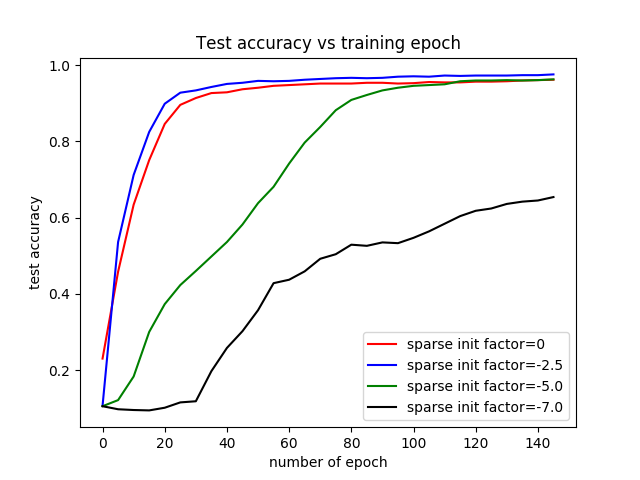}}
\subfigure[\label{fig:abla2}]{\includegraphics[width=80mm]{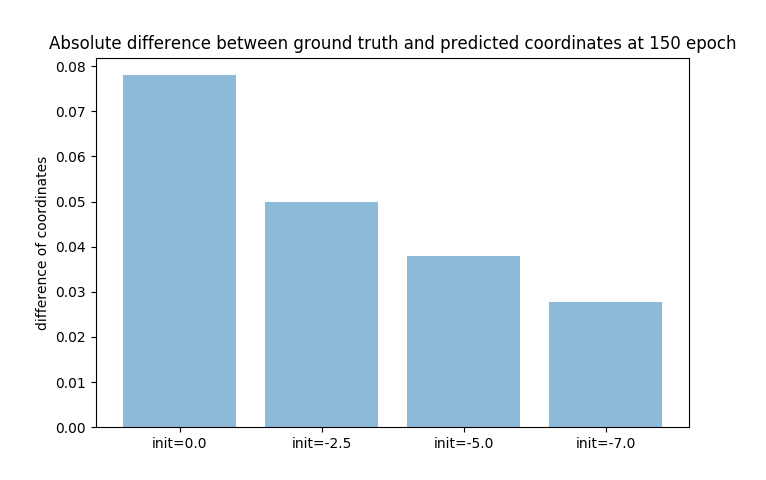}}
\caption{Figure (a): sparse initialization vs test accuracy figure. Figure (b) absolute differences of coordinates. The lower the difference is, the closer the predicted coordinates are to the ground truth coordinates. }
\end{figure}

In order to figure out how the sparse initialization influences performance of both classification and derived coordinates, we change the initial routing logit $b_{ij}$ for all the real classes. Figure~\ref{fig:abla1} shows that the lower the initial logits' values are, the slower the whole network converges. This means that the initial CapsNet~\cite{sabour2017dynamic} takes the advantage of the many active and ``usable'' capsules to achieve good performance by setting relatively high routing logits as in figure~\ref{fig:eveninit}, while these capsules may not contain distinctive factors to contribute to the capsule in the next layer. Figure ~\ref{fig:abla2} shows the performance of the weakly supervised coordinate prediction. Even though the initial $b_{ij}=-7.0$ can achieve better performance 0.027 than 0.038 we reported above with initial $b_{ij}=-5.0$, we choose the latter because figure~\ref{fig:abla1} clearly shows that these capsules with initial $b_{ij}=-7.0$ have not yet converged to achieve good classification performance. This ablation study shows that if we want the network to achieve good classification performance and learn distinctive capsule representations, a module that filters out most of the capsules (encourage sparsity) will be helpful, but it will take longer for the CapsNet to learn good capsule representation for classification. 

Other choices of sparsity encouragement include argmax and sparse entropy. Taking argmax of the capsule while ignoring the other capsule results in even less useful capsule representations in the primary capsule layer, and thus it will take more time (or even do not) converge based on the projection of figure~\ref{fig:abla1}. Sparse entropy may be a good choice and we will explore more in our future work.

\section{Benchmark model choices}\label{A:benchmark}
We have not found any work that directly produces the coordinates of objects through weakly supervised learning setting. We tried spatial transformer network\footnote{https://github.com/tensorflow/models/blob/master/research/\linebreak transformer/spatial\_transformer.py} and found that the transformation matrix of the input image is not stable enough between different trials to be a useful benchmark, mainly due to the fact that the random initialization of the transformer does not generate consistent transformation matrices for different trials. The transformed images of digits shift around and the coordinates are not as stable as CapsNet and Gradcam. 

\begin{figure}
\centering
\includegraphics[width=0.2\textwidth]{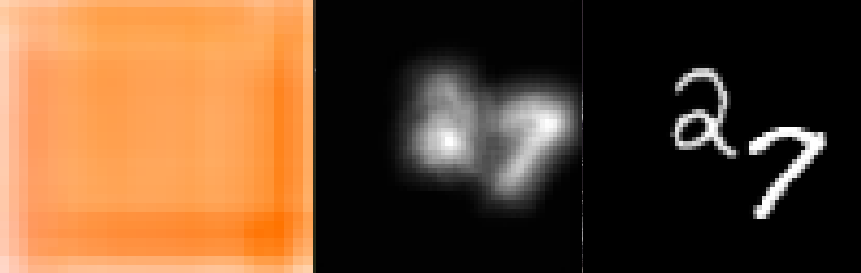}
\caption{Examples of generated flow and image compared to the ground truth image (right most). \label{fig:bench1}}
\end{figure}

Moreover, we tested a Spatio-temporal Video Autoencoder~\cite{patraucean2015spatio}, which is unsupervised and considers a sequence of frames as input that is different from our work. Without changing any of the reported hyperparameters and network structure, we can only generate the flow output as in figure~\ref{fig:bench1} which does not seem to be a fair comparison to our work. 

\end{document}